\documentclass[10pt,journal,compsoc]{IEEEtran-customize}

\ifCLASSOPTIONcompsoc
  \usepackage[nocompress]{cite}
\else
  \usepackage{cite}
\fi
\usepackage{graphicx}
\usepackage{color}
\usepackage{xcolor}
\usepackage{booktabs}
\usepackage{multirow}
\usepackage[hyphens]{url}
\usepackage{hyperref}
\usepackage{caption}
\usepackage{amsmath}
\usepackage{subcaption}
\usepackage{amsfonts, amssymb}
\usepackage{colortbl}
\usepackage{enumitem}

\definecolor{purple}{RGB}{128, 0, 128}
\definecolor{LightRed}{rgb}{1,0.92,0.92}
\definecolor{LightOrange}{rgb}{1,0.95,0.88}
\definecolor{LightYellow}{rgb}{1.0,1.0,0.84}
\definecolor{LightGreen}{rgb}{0.9,1.0,0.88}
\definecolor{LightCyan}{rgb}{0.9,1,1}
\definecolor{LightBlue}{rgb}{0.9,0.94,1}
\definecolor{LightIndigo}{rgb}{0.92,0.9,1}
\definecolor{LightMagenta}{rgb}{0.96,0.86,1}
\definecolor{DirtyWhite}{rgb}{0.96,0.96,0.96}

\DeclareSymbolFont{extraup}{U}{zavm}{m}{n}
\DeclareMathSymbol{\varheart}{\mathalpha}{extraup}{86}
\DeclareMathSymbol{\vardiamond}{\mathalpha}{extraup}{87}
\DeclareMathSymbol{\varclubsuit}{\mathalpha}{extraup}{88}

\begin{document}

\title{Text Reinforcement for Multimodal \\Time Series Forecasting}
\renewcommand\thefootnote{\fnsymbol{footnote}}
\author{
        Chen Su$^{{\spadesuit}}$, \hspace{0.1cm}
        Yuanhe Tian$^{{\varheart}}$, \hspace{0.1cm}
        Yan Song$^{{\spadesuit}}$\textsuperscript{*}, \hspace{0.1cm}
        Yongdong Zhang$^{{\spadesuit}}$
        \\
        \vspace{0.2cm}
        $^{\spadesuit}$University of Science and Technology of China \hspace{0.2cm}
        $^{\varheart}$University of Washington \\
        \vspace{0.1cm}
$^{\spadesuit}$\texttt{suchen4565@mail.ustc.edu.cn}  \hspace{0.1cm}
$^{\varheart}$\texttt{yhtian@uw.edu} \hspace{0.1cm}
\\
$^{\spadesuit}$\texttt{clksong@gmail.com} \hspace{0.1cm}
$^{\spadesuit}$\texttt{zhyd73@ustc.edu.cn}
}

\IEEEtitleabstractindextext{%

\begin{abstract}
Recent studies in time series forecasting (TSF) use multimodal inputs, such as text and historical time series data, to predict future values.
These studies mainly focus on developing advanced techniques to integrate textual information with time series data to perform the task and achieve promising results.
Meanwhile, these approaches rely on high-quality text and time series inputs, whereas in some cases, the text does not accurately or fully capture the information carried by the historical time series, which leads to unstable performance in multimodal TSF.
Therefore, it is necessary to enhance the textual content to improve the performance of multimodal TSF.
In this paper, we propose improving multimodal TSF by reinforcing the text modalities. 
We propose a text reinforcement model (TeR) to generate reinforced text that addresses potential weaknesses in the original text, then apply this reinforced text to support the multimodal TSF model's understanding of the time series, improving TSF performance.
To guide the TeR toward producing higher-quality reinforced text, we design a reinforcement learning approach that assigns rewards based on the impact of each reinforced text on the performance of the multimodal TSF model and its relevance to the TSF task. 
We optimize the TeR accordingly, so as to improve the quality of the generated reinforced text and enhance TSF performance.
Extensive experiments on a real-world benchmark dataset covering various domains demonstrate the effectiveness of our approach, which outperforms strong baselines and existing studies on the dataset.\textsuperscript{$\mathsection$}
\end{abstract}

}

\maketitle
\begingroup
  \renewcommand\thefootnote{\fnsymbol{footnote}} 
  \footnotetext[1]{Corresponding author.}
  \footnotetext[4]{the code is available at \url{https://github.com/synlp/TeR-TSF}.}
\endgroup
\IEEEdisplaynontitleabstractindextext
\IEEEpeerreviewmaketitle

\makeatletter
\def\@IEEEcompsocmakefnmark{\hbox{\normalfont\@thefnmark\ }}
\long\def\@makefntext#1{\parindent 1em\indent\hbox{\@IEEEcompsocmakefnmark}#1}
\makeatother

\makeatletter
\def\@IEEEcompsocmakefnmark{\hbox{\normalfont\@thefnmark.\ }}
\long\def\@makefntext#1{\parindent 1em\indent\hbox{\@IEEEcompsocmakefnmark}#1}
\makeatother

\renewcommand{\thefootnote}{\arabic{footnote}}

\section{Introduction}
\label{introduction}

Time series forecasting (TSF) aims to predict future values based on historical time series and plays a crucial role in decision-making across various scenarios.
It typically analyzes temporal patterns, trends, and seasonal variations to model dependencies between past and future data points.
The ability to automatically predict future values enables one to optimize resources, mitigate risks, and enhance operational efficiency.
Thus, TSF applications span diverse domains, such as finance for stock price predictions \cite{tsf4stock1, tsf4stock2, tsf4stock3}, meteorology for weather forecasting \cite{tsf4weather1, tsf4weather2, tsf4weather3}, energy management for demand estimation \cite{tsf4energy1, tsf4energy2, tsf4energy3}, supply chain optimization for inventory planning \cite{tsf4supply1, tsf4supply2, tsf4supply3}.

Early research on TSF that exclusively utilizes historical numerical time series for prediction has already shown promising performance \cite{logtrans, informer, autoformer, pyraformer, fedformer, etsformer,su2025diffusion}. 
These studies enhance prediction performance by mining temporal patterns such as periodic and multi-resolution components within historical numerical sequences. 
However, such time series analysis has encountered a bottleneck owing to the inherent limitations of information contained in historical numerical observations, and thus may require other modalities of data to deliver critical predictive insights. 
Therefore, the integration of multimodal data into TSF attracts increasing attention because of its potential to capture richer contextual information than unimodal approaches that rely solely on historical data points \cite{time-mmd, contextformer, hybrid-mmf}.
Among all types of multimodal data, many studies utilize textual inputs (such as expert analyses and public sentiment) to enhance time series modeling, which provides complementary information that enables models to interpret abrupt fluctuations or non-recurring anomalies triggered by external factors.
To effectively leverage such cross-modal relationships, advanced fusion strategies are required to align temporal and text dependencies. 
In doing so, early approaches \cite{multimodalTSFGNN1, multimodalTSFGNN2} fuse numerical and textual features through graph attention mechanisms.
Subsequent studies \cite{contextformer, textfusionhts} employ Transformer-based cross-attention mechanisms to enable adaptive cross-modal interactions, where noisy or sparse textual input remarkably hurts their performance.
To better encode textual and time series information, recent studies \cite{time-llm,chattime,su2025fusing} utilize large language models (LLMs), which are proven to be effective in achieving outstanding performance in multimodal content processing.
However, these approaches treat time-series data as fixed inputs, omitting their nature of stochastic fluctuations.
Following research \cite{mcd-tsf} consequently integrates probabilistic frameworks (e.g., diffusion models) to explicitly model time-series dynamics while retaining LLMs' textual processing strengths, thereby systematically addressing stochastic uncertainties.
These multimodal TSF approaches achieve remarkable performance and demonstrate the effectiveness of utilizing textual information.
Still, critical challenges persist in real-world deployment scenarios where the textual modality suffers from quality issues such as incompleteness and misalignment with time series data \cite{obst2019textual, time-mmd, hybrid-mmf}.
Specifically, the issue of missing textual data at certain times significantly affects TSF when it is modeled relying on multimodal input.
Besides, misaligned texts (e.g., out-of-date news) incorrectly linked to current data points at some times (e.g., stock prices) provide irrelevant signals that mislead the model in understanding the inherent temporal dependencies in the time series, even leading multimodal models to obtain worse performance than unimodal baselines.
An effective solution to address the data quality issues is to perform data reinforcement, which augments original data and has been demonstrated to be effective for many tasks \cite{data_aug_survey1, data_aug_survey2, data_aug_survey3, data_aug_survey4, data_aug_survey5}.
Particularly, for time series modeling,
conventional approaches \cite{data_aug_ts1, data_aug_ts2, data_aug_ts3} such as interpolation and noise injection, alongside generative models, already demonstrate their effectiveness in generating augmented time series to address the data scarcity or class imbalance issues.
They are restricted in focusing on augmenting unimodal time series and rely on static augmentation strategies, making them inapplicable to generating text with a dynamic relationship with numerical observations.
Recent studies \cite{timecap} employ LLMs, leveraging their pre-trained knowledge and powerful reasoning capabilities, to adaptively generate corresponding text based on numerical observations.
Nevertheless, they cannot directly determine how the generated text assists downstream TSF, making it difficult to obtain direct feedback on the quality of generated texts, which in turn makes it challenging to apply supervised fine-tuning approaches to optimize the LLMs.
While directly assessing the quality of generated text remains challenging, there are several indirect evaluation metrics 
(e.g., the performance on downstream tasks and the language used to describe the time series data),
which are potentially rewards to optimize LLMs through reinforcement learning (RL).
Thus, it is expected to advance multimodal TSF with data augmentation optimized by RL and LLMs, so that high-quality texts are generated and aligned with numerical observations to improve TSF.

In this paper, we propose a novel RL-driven data augmentation framework, TeR-TSF, to address the pervasive challenges of textual data absence and misalignment in multimodal TSF. 
The objective is to generate semantically coherent and temporally aligned reinforced text from raw multimodal inputs,  so that enhancing prediction accuracy without reliance on fully or perfectly aligned auxiliary texts.
TeR-TSF utilizes an LLM to generate contextually relevant text based on both historical time-series data and available original texts.
The generated texts, equally described as reinforced texts, are integrated with historical time series to form new inputs to the multimodal TSF model, which produces forecasts for future time series values. 
Particularly for RL training, a dual-objective rewards evaluate text generation quality through three synergistic components.
The first component measures text-enhanced prediction accuracy by computing the negative distance between model-generated and ground-truth time-series values.
The second evaluates domain relevance by statistically analyzing the occurrence density of task-specific keywords in the generated text, verifying its adherence to descriptive requirements.
{
These rewards jointly identify high-quality reinforced text as positive examples and suboptimal ones as negative examples, forming a dynamic training signal that enables direct preference optimization (DPO) \cite{dpo} to iteratively refine the LLM’s generation.
By aligning the LLM with both numerical forecasting objectives and cross-modal semantic consistency, our approach generates text that not only compensates for missing or noisy real-world text but also disentangles spurious temporal-textual correlations. 
Comprehensive experiments on a benchmark dataset demonstrate the effectiveness of our approach, which consistently outperforms state-of-the-art multimodal TSF baselines.
Further analyses confirm that the RL-driven reinforcement mechanism contributes to the improvement in forecasting accuracy.
These results validate the effectiveness of our approach in reinforcing incomplete and misaligned text into a comprehensive summary for robust TSF, providing new solutions for real-world deployment where textual modalities are inherently imperfect.
}

\begin{figure*}[t]
\begin{center}
\centerline{\includegraphics[width=2\columnwidth, trim=0 10 0 0]{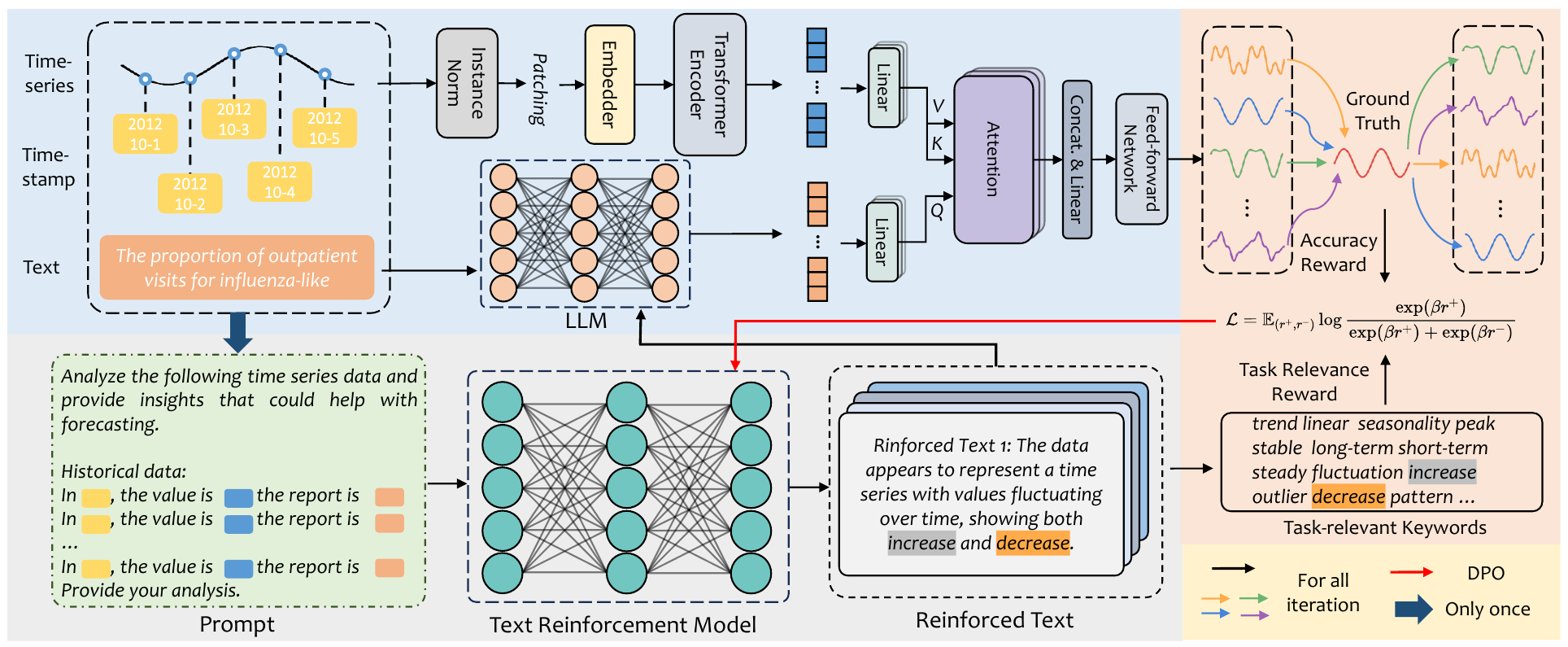}}
\caption{
{
An illustration of the TeR-TSF Model. The left-top panel depicts the multimodal TSF model, which separately encodes time-series data and text, fuses the multimodal features, and produces predictions. The left-bottom panel illustrates the process of structuring multimodal time-series data into a prompt for an LLM to generate reinforced text. The right panel explains the construction of preference data using the model predictions and the reinforced text for DPO training.
}
}
\label{Overall Architecture}
\end{center}
\vskip -0.3in
\end{figure*}

\section{Related Work}

\subsection{Text Helps TSF}

Multimodal TSF with text \cite{xu2018stock, wang2024news, liu2025timecma, liu2024can} needs to extract textual features and integrate them with time series features through multimodal fusion, where the resulting fused information is used to predict future time series.
Early approaches \cite{xian2021multi, wang2023forecasting} predominantly rely on convolutional or Transformer-based architectures to extract semantic representations from text, yet often struggle with capturing nuanced text representation.
Subsequent studies \cite{time-mmd, contextformer, textfusionhts, gpt4mts, unitime} leverage LLMs for better text representation and combines textual embeddings with temporal features through relation graphs or cross-attention mechanisms. 
For example, Liu et al. (2024) \cite{time-mmd} propose fixed-weight summation between text-derived predictions and historical numerical observation-based predictions.
Wang et al. (2024) \cite{chattime} convert numerical time series into pseudo-textual tokens for LLM-based question-answering frameworks to extract predictions.
Zhou et al. (2025) \cite{textfusionhts} develop an adaptive feature fusion using cross-attention layers before final prediction through feedforward networks. 
{
These approaches primarily focus on improving text representations and text-time series feature fusion without further processing of the text to obtain its key information.
Different from existing studies, we enhance multimodal fusion by reinforcing textual content, thereby improving multimodal TSF.
}

\subsection{TSF with Reinforcement Learning}

Reinforcement Learning \cite{wang2024inspiration, hu2024review, tang2025deep, tian2024learning, kumar2024training} guides agents to dynamically interact with their environment through rewarding, iteratively optimizing decision-making policies to accomplish complex task objectives.
Recent studies \cite{shaik2024graph, niu2025langtime, fu2022reinforcement, perepu2020reinforcement, qi2025timehf} that explore the application of RL in TSF primarily focus on model-level optimization to enhance forecasting performance.
For example, Fu et al. (2022) propose a deep deterministic policy gradient (DDPG) framework for dynamic model ensemble adjustment, where the policy network takes temporal features and historical prediction errors as state inputs, and then predicts adaptive weights of base models to forecast non-stationary data.
Niu et al. (2025) \cite{niu2025langtime} employ proximal policy optimization (PPO) to refine autoregressive forecasting trajectories, mitigating error accumulation in multi-step predictions and improving cross-domain generalization.
Qi et al. (2025) \cite{qi2025timehf} integrate expert knowledge via RL from human feedback (RLHF) to optimize predictions in complex scenarios, boosting zero-shot performance accordingly.
Another line of studies \cite{liu2022synthetic, yuan2024learning} that applies RL to this task is data augmentation, thereby improving TSF performance.
For instance, Yuan et al. \cite{yuan2024learning} design variance-based rewards to guide variational masked autoencoders (VMAE) in generating synthetic data that addresses model overfitting, thereby enhancing the generalization capability of the model.
These approaches mainly focus on augmenting time series data, with less attention paid to enhancing different modalities for the task. 
Different from existing studies, we propose an RL-based approach to reinforce text for better multimodal TSF.
\subsection{Data Augmentation for TSF}

Time series data augmentation aims to enhance model performance and generalization capabilities by generating additional high-quality instances. 
Early approaches employ pre-processing such as noise injection, scaling, and shifting for data expansion \cite{TimeSeriesAugSurvey1, TimeSeriesAugSurvey2, TimeSeriesAugConv1, TimeSeriesAugConv2, TimeSeriesAugConv3, TimeSeriesAugConv4}, which are straightforward to implement but limited in data diversity and prone to damaging the inherent structure and temporal dependencies of time series data.
Subsequent studies \cite{TimeSeriesAugFreq1, TimeSeriesAugFreq2, TimeSeriesAugFreq3} develop signal processing techniques that attempt to preserve statistical characteristics and temporal patterns while expanding original data.
{
For example, Forestier et al. \cite{TimeSeriesAugDTW} leverage dynamic time warping (DTW) to expand sparse datasets by generating synthetic samples through weighted averaging. 
Gao et al. \cite{TimeSeriesAugFreq1} enhance the generalization capability of anomaly detection models by generating diversified data through amplitude and phase perturbations in the frequency domain.
However, these approaches rely heavily on predefined and manually created rules and prior knowledge of data characteristics.
The advent of deep learning introduces data-driven data augmentation paradigms \cite{TimeSeriesAugVAE1, TimeSeriesAugVAE2, TimeSeriesAugVAE3, TimeSeriesAugVAE4} that are capable of automatically learning sophisticated temporal patterns.
For instance, Goubeaud et al. \cite{TimeSeriesAugVAE4} enhance sparse training sets and improve time series modeling by training separate variational autoencoders for each category to generate synthetic time series data, thereby enhancing the model performance.
Huang et al. \cite{TimeSeriesAugGAN} encode time series into images that preserve temporal dependencies, utilizing a deep convolutional generative adversarial network to generate synthetic data for filling in missing values.
Concurrently, RL-based approaches emerge. 
For instance, Liu et al. \cite{liu2022synthetic} generate synthetic financial data using a GAN and employ it as a training environment to enhance the trading strategies of deep reinforcement learning (DRL) agents. 
This approach addresses the scarcity of real-world data while improving the profitability and robustness of models in financial trading.
While these studies demonstrate notable success in unimodal time series augmentation, their direct application to multimodal time series data remains challenging due to the essential requirement for preserving inter-modal alignment.
Different from existing approaches that primarily address unimodal scenarios, this work specifically targets the reinforcement of multimodal time series data with particular emphasis on textual modality. The reinforced text ensures the cross-modal alignment with time series, thus improving the performance of TSF tasks.
}

\section{The Approach}

Given input time series $\mathcal{S}$ across timestamps $\{t_0,t_1,...,t_L\}$ and their available accompanied texts $\mathcal{E}$, our approach improves a multimodal TSF model by reinforcing $\mathcal{E}$ with {the time series $\mathcal{S}$}
and predicts the time-series values $\mathcal{Y}=\{t_L,t_{L+1},...,t_{L+H}\}$ in the future $H$-steps.
The overall architecture of our approach is illustrated in Fig. \ref{Overall Architecture}.
Specifically, our approach firstly utilizes {an LLM (which serves as the text reinforcement model (TeR))} 
to process the $\mathcal{S}$ and $\mathcal{E}$ to generate the reinforced text $\mathcal{E}_{\text{aug}}$.
{
Then, a multimodal TSF model $f_{\text{TSF}}$ trained on multimodal data $(\mathcal{S},\mathcal{E})$ integrates both time series $\mathcal{S}$ and the generated $\mathcal{E}_{\text{aug}}$ to predict future time-series values $\mathcal{Y}$.
}
Finally, an RL process is utilized to optimize the $\mathcal{E}_{\text{aug}}$ generation based on different rewards that assess the contributions of $\mathcal{E}_{\text{aug}}$ to the TSF task.
The details of the processes are presented in the following text.

\subsection{Text Reinforcement}
In generating reinforced text based on the time series $\mathcal{S}$ with its original accompanying text $\mathcal{E}$, we utilize a pre-trained LLM to do so.
Consider that $\mathcal{S}$ is numeric data and hard to be understood by an LLM primarily trained on texts \cite{AreLLMUseful, chronos},
we translate the time series data $S$ into text $\mathcal{S}_\text{txt}$ to make it easier to fit the LLM to better generate the reinforced text.
{Inspired by Chattime \cite{chattime}, we convert $\mathcal{S}$ into fixed 4-level precision. This approach avoids introducing additional tokens to represent unnecessary precision levels, since LLMs do not need to perform complex computational tasks on the numerical data.}
{To enhance LLMs' comprehension of the overall characteristics of time series, we directly supplement them with descriptive information about the time series.}
In doing so, the auxiliary descriptive text $\mathcal{A}_\text{txt}$ is systematically constructed by extracting statistical features (including mean, variance, etc.) and incorporating metadata attributes (such as timestamp, sampling frequency, etc.) derived from $\mathcal{S}$.
{%
For example, for a time series such as \textit{[0.12931496, 0.38298721, 0.31423577]}, we convert it into the text ``\textit{0.1293 0.3829 0.3142}'' (remember that we only keep the four digits after the decimal point).
This is followed by the descriptive text $\mathcal{A}_\text{txt}$: ``\textit{Mean: 0.2754, Variance: 0.0114}'' (i.e., the descriptive text).
}
Therefore, the $\mathcal{S}_\text{txt}$ and the $\mathcal{A}_\text{txt}$ together describe the complete information of $\mathcal{S}$.
Finally, the $\mathcal{S}_\text{txt}$ and $\mathcal{A}_\text{txt}$ are concatenated with the original text $\mathcal{E}$ and task-specific prompts $\mathcal{P}$ (for example, ``\textit{Making predictions based on the information above.}'') to form an input for the LLM to generate reinforced text $\mathcal{E}_\text{aug}$ through
\begin{equation}
    \mathcal{E}_{\text{aug}}=f_{\text{TeR}}(\mathcal{S}_{\text{txt}}, \mathcal{A}_{\text{txt}},\mathcal{E},\mathcal{P})
\end{equation}
During this process, it is expected that the LLM is able to improve the quality of $\mathcal{E}$ with $\mathcal{E}_\text{aug}$, and accordingly help a multimodal TSF model to better understand the time series and thus improve its performance.

\subsection{Multimodal TSF with Reinforced Text}
{
The multimodal TSF model integrates time series $\mathcal{S}$ and reinforced text $\mathcal{E}_{\text{aug}}$ to generate future predictions $\widehat{\mathcal{Y}}$ by leveraging both temporal patterns and semantically textual features. 
These cross-modal inputs are fused at the feature level through the multimodal TSF model to capture intricate interactions between numerical temporal patterns and textual content, which is formulated as:
\begin{equation}
    \widehat{\mathcal{Y}}=f_{\text{TSF}}(\mathcal{S}, \mathcal{E}_{\text{aug}})
\end{equation}
Specifically, within our approach, building upon Zhou et al. \cite{textfusionhts}, we leverage a patch-wise PatchTST model to encode the time series data $\mathcal{S}$. 
Concurrently, a LLM processes the augmented text $\mathcal{E}_{\text{aug}}$ for tokenization and embedding. 
For the time series, the initial step partitions the normalized data into patches. 
These patch tokens undergo embedding and are subsequently processed by a multi-layer Transformer Encoder to capture temporal dependencies by
\begin{equation}
    \mathbf{S} = \operatorname{PatchTST}(\mathcal{S})
\end{equation}
where $\mathbf{S}$ represents the encoded representation of the time series $\mathcal{S}$ produced by the PatchTST model. 
For the augmented text $\mathcal{E}_{\text{aug}}$, the process firstly tokenizes the text and next utilizes the LLM embedding layer to
convert the resulting tokens into vector representations, which are denoted by $\{\mathbf{e}_i \mid i=1,2,\dots,n\}$ with $n$ representing the total number of text tokens.
We aggregate the token representations to compute the text representation vector $\mathbf{e}$ by
\begin{equation}
    \mathbf{e} = \frac{1}{n} \sum_{i=1}^{n} \mathbf{e}_{i}
\end{equation}
Then, we utilize $\mathbf{e}$ as a query and the encoded time series $\mathbf{S}$ as both keys and values within a cross multi-head attention mechanism.
This operation generates a fused semantic vector integrating information from both modalities by
\begin{equation}
    \mathbf{z} = \operatorname{CrossAttention}(\mathbf{e}, \mathbf{S}, \mathbf{S})
\end{equation}
where the resulting vector $\mathbf{z}$ embodies the combined semantics extracted from the time series and text data. 
Finally, this fused vector $\mathbf{z}$ passes through a feed-forward network layer. 
The output of this layer constitutes the prediction for the future time series values $\widehat{\mathcal{Y}}$. 
This multimodal time series forecasting architecture directly and effectively fuses time series and textual modal data to generate predictions. 
These predictions $\widehat{\mathcal{Y}}$ serve as the basis for the subsequent evaluation of the generated text quality.
}

\subsection{Optimization with Reinforcement Learning}

{
To optimize the LLM for generating better reinforced texts $\mathcal{E}_{\text{aug}}$, we design a reward function to measure the quality of different reinforced texts and then optimize the LLM through DPO.
The details are illustrated as follows.
}

\subsubsection{Reward Computation}
{
The reward generator (RG) in our approach computes the reward $r$ by evaluating the quality of generated reinforced texts based on the reinforced text $\mathcal{E}_{\text{aug}}$ itself and the prediction $\widehat{\mathcal{Y}}$ produced by the multimodal TSF model.
To comprehensively evaluate the quality of the reinforced text from various aspects, we design a dual reward framework to facilitate the selection of generated texts that best approximate the ideal characteristics in practical applications.
}

\textbf{{Reward 1: Prediction Accuracy Reward}}
{
The first reward $r_1$ aims to evaluate whether the reinforced texts effectively assist the TSF model in improving its forecasting performance. 
Directly evaluating the explicit contribution of text to time-series prediction is inherently challenging due to the implicit and abstract nature of their relationship.
To address this, we leverage the mean squared error (MSE) between the predicted future time-series $\widehat{{\mathcal{Y}}}$ generated by the TSF model and the ground truth $\mathcal{Y}$. 
The reward $r_1$ is formulated as the negative MSE to encourage the TeR to produce texts that minimize forecasting errors by
\begin{equation}
     r_1 = -\frac{1}{H} \sum_{i=1}^H (\mathcal{Y}_i - \widehat{\mathcal{Y}}_i)^2
\end{equation}
where $H$ is the number of predicted timesteps. 
This design indirectly aligns the textual content with the temporal dynamics of the data based on the output of the TSF model.}

\textbf{{Reward 2: Task Relevance Reward}}
{
The third reward term $r_2$ enforces lexical constraints to ensure that generated texts adhere to domain-specific terminology and remain interpretable to humans. 
We predefined a set of task-relevant keywords (such as ``\textit{peak}'', ``\textit{fluctuation}'', ``\textit{seasonality}'', etc.) based on domain expertise. 
A binary indicator function $\mathbb{I}(w \in \mathcal{K})$ checks the presence of keywords $w$ from the keyword set $\mathcal{K}$ in the generated text. 
The reward $r_2$ is calculated as the normalized count of such keywords:
\begin{equation}
     r_2 = \frac{\sum_{w \in \mathcal{W}} \mathbb{I}(w \in \mathcal{K})}{|\mathcal{K}|}
\end{equation}
Beyond improving linguistic relevance, this reward facilitates the generation of human-readable explanations that reveal underlying temporal characteristics.
}

{
The final reward $r$ is obtained by the weighted sum of two reward terms through $r = w_1 \cdot r_1+w_2 \cdot r_2$.
This dual reward ensures that the proposed TeR jointly optimizes forecasting utility and task relevance.
}

\subsubsection{Training with DPO}
{
To enhance the quality of reinforced text generated by the TeR, we design a reward-guided refinement approach using DPO. 
The core idea is to iteratively adjust TeR’s generation strategy to prioritize high-quality reinforced text, guided by the reward. 
This refinement process contains three key stages: candidate generation, reward-based ranking, and optimization via DPO, executed cyclically over $m$ rounds.
Given input text $\mathcal{S}$ and its original text $\mathcal{E}$, TeR generates $k$ diverse enhanced candidates $\{\mathcal{E}_{\text{aug}}^1, \mathcal{E}_{\text{aug}}^2, \ldots, \mathcal{E}_{\text{aug}}^k\}$. 
This step aims to explore variations in the generation space, thereby facilitating the process to find the reinforced text that best matches the reward metric. 
Next, the reward generator model assigns a scalar score $r^i$ to each candidate $\mathcal{E}_{\text{aug}}^i$, reflecting its comprehensive evaluation on the multi-dimensional reward metric.
Then, candidates are ranked by their reward scores, with the highest-ranked candidate $\mathcal{E}_{\text{aug}}^+$ selected as the positive example and the lowest-ranked $\mathcal{E}_{\text{aug}}^-$ as the negative example, forming a preference pair for DPO training. 
The DPO optimization process minimizes a loss function that amplifies the likelihood difference between preferred and dispreferred outputs under the current policy, effectively steering TeR's generation strategy toward high-reward regions of the text space. 
Specifically, to refine TeR, we minimize the DPO loss $\mathcal{L}$ computed by
\begin{equation}
    \mathcal{L} = \mathbb{E}_{(r^+, r^-)} \log \frac{\exp(\beta r^+)}{\exp(\beta r^+) + \exp(\beta r^-)}
\end{equation}
where $r^+$ and $r^-$ correspond to the rewards of positive examples and negative examples, respectively, and $\beta$ is a temperature parameter scaling the reward difference. 
The objective strengthens TeR’s preference for high-reward candidates by widening the likelihood gap between $\mathcal{E}_{\text{aug}}^+$ and $\mathcal{E}_{\text{aug}}^-$. 
After each optimization iteration, the updated TeR regenerates a new set of candidates for the subsequent round of reward evaluation and preference-based training, creating a closed-loop refinement cycle. 
{
Meanwhile, the multimodal TSF model is trained on the multimodal data $(\mathcal{S},\mathcal{E}_{\text{aug}}^+)$ reconstructed using the reinforced text generated in the previous iteration, in order to compute the current iteration's prediction accuracy reward.
}
Through $m$ iterations, TeR progressively refines its text generation strategy by building upon previous optimizations. This enables the average reward score of generated reinforced text candidates to incrementally improve, ultimately allowing the TeR to consistently produce high-quality, readable text that effectively aligns with temporal patterns.
In summary, through this self-improving framework, the proposed approach achieves sustained enhancements in text reinforcement quality without requiring external human annotations.}

\section{Experiment Settings}
\subsection{Datasets}
\begin{table}[t]
    \centering
    \caption{{
Statistical overview of the Time-MMD dataset across eight domains. For each domain, the recording frequency of the time series is specified. The ``Time Steps'' indicates the total length of the aggregated time series data per domain. The ``Text Number'' represents the total count of textual records associated with each domain.
    }
    \label{tab: detail time-mmd dataset}
    }
    \begin{tabular}{l|rrr}
    \toprule
        \textbf{Domains} &  \textbf{Frequency} & \textbf{Time Steps} & \textbf{Text Number} \\ 
    \midrule
        Agriculture & Monthly & 496 & 890 \\ 
        Climate & Weekly & 496 & 582 \\ 
        Economy & Monthly & 432 & 435 \\
        Energy & Weekly & 1,479 & 354 \\
        Environment & Daily & 11,102 & 141 \\
        Health & Weekly & 1,389 & 489 \\
        SocialGood & Monthly & 900 & 347 \\
        Traffic & Monthly & 531 & 367 \\
    \bottomrule
    \end{tabular}
\end{table}
{
Following existing studies, our experiments are conducted on the Time-MMD dataset \cite{time-mmd}, which incorporates both time series and texts and is specifically designed to evaluate model performance in multimodal TSF scenarios.
Specifically,
Time-MMD spans eight domains, namely, \textit{Agriculture}, \textit{Climate}, \textit{Economy}, \textit{Energy}, \textit{Environment}, \textit{Health}, \textit{Social Good}, and \textit{Traffic}, featuring diverse time series features and corresponding text.
However, the Time-MMD dataset presents challenges related to text sparsity and text irrelevance.
Text sparsity refers to the shortage of textual data in the domain, which frequently results in historical windows of time series forecasting containing no text at all.
Text irrelevance manifests as recorded textual content failing to correspond meaningfully to the target variables described by the time series. 
Therefore, this dataset serves as a good resource to evaluate the proposed framework's capability in handling sparse and mismatched textual information.
For each domain, all time series are normalized to zero mean and unit standard deviation to mitigate scale variations across domains.}
Following previous studies \cite{autoformer, fedformer, Non-stationary-transformers, crossformer, time-llm, mcd-tsf}, we chronologically split each dataset into training, validation, and test sets with a 7:1:2 ratio to preserve temporal dependencies and simulate real-world forecasting scenarios, ensuring no future data leaks into earlier splits.

\subsection{Baselines and Comparing Approaches}

To evaluate our proposed TeR-TSF, we compare it with four baseline models outlined below:
\begin{itemize}[leftmargin=1em]
    \item \textbf{TSF-only}: {The TSF model in TeR-TSF without text input, i.e., a standard unimodal TSF model.} 
    \item \textbf{TSF+Text}: {The TSF model in TeR-TSF where the texts are used as a part of the input, i.e., TFHTS model \cite{textfusionhts}.} 
    \item \textbf{TSF+TeR}: This model extends the TSF model by incorporating the TeR to augment multimodal data.
    TeR herein is not trained with reinforcement learning.
    \item \textbf{TSF+TeR+r1}: This model enhances TeR-TSF with the {RG} model using Reward-1 and trains the TeR via RL.
    \item \textbf{TSF+TeR+r12}: This model further improves TeR-TSF by introducing the {RG} model with combined Reward-1 and Reward-2, with the TeR trained via RL.
\end{itemize}
To further demonstrate the effectiveness and superiority of our TeR-TSF, we extensively compare it against existing state-of-the-art multimodal TSF models. 
{These models are divided into three groups according to their model types.
Specifically, these models include:
\begin{itemize}[leftmargin=1em]
    \item Transformer-based models: \textbf{MM-TSF} \cite{time-mmd}, \textbf{TFHTS} \cite{textfusionhts}.
    \item LLM-based models: \textbf{Time-LLM} \cite{time-llm}.
    \item Diffusion-based models: \textbf{MCD-TSF} \cite{mcd-tsf}.
\end{itemize}
}

\subsection{Implementation Details}

According to MCD-TSF \cite{mcd-tsf}, the historical and forecast lengths used in Time-MMD are frequency-dependent: monthly-frequency series employ a 36-step historical window with forecast horizons of 6, 12, and 18 steps; weekly-frequency series utilize a 96-step historical length paired with 12, 24, and 36-step predictions; while daily-frequency series adopt an extended 336-step historical length for 48, 96, and 192-step forecasting. 
Besides, textual inputs synchronize with their time series counterparts by employing identical historical length, with all available textual data within these ranges being incorporated for analysis. 

To generate high-quality reinforced text, it is essential to use an advanced text generator since the text modeling capability of a model is critical for generating informative and fluent texts \cite{pennington2014glove,ijcai2018-607,han2018hyperdoc2vec,brown2020language,dubey2024llama}.
Therefore, for the architecture of the TeR model, we use Qwen3-1.7B\footnote{\url{https://huggingface.co/Qwen/Qwen3-1.7B}} \cite{yang2025qwen3}, which is the latest LLM of the Qwen series with outstanding performance on many text processing tasks \cite{warner2024smarter,chen2025towards,liu2025balanced,tian2025large}.
For the multimodal TSF model, we utilize the TextFusionHTS framework \cite{textfusionhts}.
All models are configured with their default hyper-parameter settings.
Specifically, we employ the default tokenizer of Qwen3-1.7B to tokenize the text. A 28-layer Qwen3-1.7B model then generates the reinforced text.
The TSF model uses PatchTST \cite{patchtst} to encode the time series and encodes textual inputs with the tokenizer and embedding layer from Llama-3.1 8B \cite{dubey2024llama}.

For hyperparameter settings, we set the default value of the number of training rounds and the number of generated reinforced texts per round to $4$ and $2$, respectively. 
During DPO, we employ the low-rank adaptation (LoRA) \cite{lora} technique for efficient parameter fine-tuning. 
The learning rate, LoRA rank, preference optimization beta, and preference loss function are configured with default values of $0.00005$, $8$, $0.1$, and sigmoid, respectively. 
A cosine learning rate scheduler is used with a warmup ratio of 0.1.

{
Following existing studies \cite{patchtst,dlinear}, we employ Mean Squared Error (MSE) and Mean Absolute Error (MAE) as evaluation metrics, where lower values indicate better predictive performance. The reported results represent the average measurements across all forecasting horizons.
}

\begin{table*}[t]
\centering
\caption{{Performance evaluation in terms of MSE and MAE for baseline models and TeR-TSF (i.e., TSF+TeR+r12). The ``Avg.'' row shows the average across all domains on the test sets. The best and second-best results (lower is better) are denoted by \textbf{boldface} and \underline{underlines}, respectively.}}
\label{tab:baselines}
\begin{tabular}{c|c|cccccccccc}
\toprule
\multicolumn{2}{c|}{} & \multicolumn{2}{c}{\textbf{TSF-only}} & \multicolumn{2}{c}{\textbf{TSF+Text}} & \multicolumn{2}{c}{\textbf{TSF+TeR}} & \multicolumn{2}{c}{\textbf{TSF+TeR+r1}} & \multicolumn{2}{c}{\textbf{TSF+TeR+r12}} \\
\multicolumn{2}{c|}{}                           & MSE        & MAE       & MSE          & MAE         & MSE             & MAE      & MSE             & MAE  & MSE             & MAE \\
\midrule
\multicolumn{2}{l|}{Agriculture}      & 0.441      & 0.492   & 0.571    & 0.564   & 0.577        & 0.557       & \underline{0.388}      & \underline{0.427}       & \textbf{0.338}       & \textbf{0.402}         \\
\multicolumn{2}{l|}{Climate}          & 1.657      & 1.015   & 1.782    & 0.948   & 2.972        & 1.406       & \underline{1.472}      & \underline{0.946}       & \textbf{1.348}       & \textbf{0.874}         \\
\multicolumn{2}{l|}{Economy}          & 0.376      & 0.497   & 0.278    & 0.442   & 0.280        & 0.668       & \underline{0.253}      & \underline{0.438}       & \textbf{0.240}       & \textbf{0.387}         \\
\multicolumn{2}{l|}{Energy}           & 0.250      & 0.398   & 0.290    & 0.403   & 0.304        & 0.414       & \underline{0.234}      & \underline{0.344}       & \textbf{0.202}       & \textbf{0.324}         \\
\multicolumn{2}{l|}{Environment}      & 0.254      & 0.364   & 0.262    & 0.368   & 0.278        & 0.380       & \underline{0.253}      & \underline{0.354}       & \textbf{0.251}       & \textbf{0.346}         \\
\multicolumn{2}{l|}{Health}           & 1.716      & 0.881   & 1.514    & 0.809   & 1.804        & 0.878       & \underline{1.504}      & \underline{0.778}       & \textbf{1.384}       & \textbf{0.756}         \\
\multicolumn{2}{l|}{SocialGood}       & 1.362      & 0.689   & 1.352    & 0.677   & 1.316        & 0.692       & \underline{1.316}      & \underline{0.671}       & \textbf{1.199}       & \textbf{0.587}         \\
\multicolumn{2}{l|}{Traffic}          & 0.119      & 0.192   & 0.101    & 0.189   & 0.100        & 0.187       & \underline{0.099}      & \underline{0.187}       & \textbf{0.092}       & \textbf{0.154}         \\
\multicolumn{2}{l|}{Avg.}             & 0.771      & 0.566   & 0.768    & 0.550   & 0.953        & 0.647       & \underline{0.689}      & \underline{0.518}       & \textbf{0.631}       & \textbf{0.478}         \\
\bottomrule
\end{tabular}%
\end{table*}

\begin{table*}[]
\centering
\caption{
{
Performance comparison of existing studies using MSE and MAE metrics. The "Avg." row presents the average performance across multiple domains on the test sets. \textbf{Boldface} and \underline{underlines} denote the best and second-best results, respectively (lower values indicate better performance).
}
}
\label{tab:comp_sota}
\begin{tabular}{cc|cccccccccc}
\toprule
\multicolumn{2}{c|}{} & \multicolumn{2}{c}{\textbf{TFHTS}} &  \multicolumn{2}{c}{\textbf{MCD-TSF}} & \multicolumn{2}{c}{\textbf{Time-LLM}} & \multicolumn{2}{c}{\textbf{MM-TSF}} & \multicolumn{2}{c}{\textbf{TeR-TSF}} \\
\multicolumn{2}{c|}{}     & MSE             & MAE         & MSE          & MAE          & MSE               & MAE               & MSE             & MAE            & MSE            & MAE            \\
\midrule
\multicolumn{2}{l|}{Agriculture}      & 0.571    & 0.564      & \textbf{0.222}   & \textbf{0.322}   & 0.448    & 0.491   & 0.714   & 0.607    & \underline{0.338}   & \underline{0.402}   \\
\multicolumn{2}{l|}{Climate}          & 1.782    & 0.948      & \underline{1.583}   & 0.971   & 1.775    & \underline{0.921}   & 1.278   & 1.041    & \textbf{1.348}   & \textbf{0.874}   \\
\multicolumn{2}{l|}{Economy}          & 0.278    & 0.442      & \underline{0.249}   & \textbf{0.374}   & 0.303    & 0.397   & 1.103   & 0.820    & \textbf{0.240}   & \underline{0.387}   \\
\multicolumn{2}{l|}{Energy}           & 0.290    & 0.403      & \textbf{0.153}   & \textbf{0.293}   & 0.234    & 0.336   & 0.610   & 0.601    & \underline{0.202}   & \underline{0.324}   \\
\multicolumn{2}{l|}{Environment}      & \underline{0.262}    & \underline{0.368}      & 0.275   & 0.379   & 0.283    & 0.398   & 0.336   & 0.467    & \textbf{0.251}   & \textbf{0.346}  \\
\multicolumn{2}{l|}{Health}           & 1.514    & \underline{0.809}      & \underline{1.496}   & 0.811   & 1.551    & 0.818   & 1.368   & 0.760    & \textbf{1.384}   & \textbf{0.756}   \\
\multicolumn{2}{l|}{SocialGood}       & 1.352    & 0.677      & \textbf{1.035}   & \textbf{0.569}   & 1.047    & 0.631   & 1.067   & 0.598    & \underline{1.199}   & \underline{0.587}   \\
\multicolumn{2}{l|}{Traffic}          & 0.101    & 0.189      & \underline{0.093}   & \underline{0.159}   & 0.182    & 0.227   & 0.225   & 0.402    & \textbf{0.092}   & \textbf{0.154}   \\
\multicolumn{2}{l|}{Avg.}             & 0.768    & 0.550      & \underline{0.638}   & \underline{0.484}   & 0.727    & 0.527   & 0.837   & 0.662    & \textbf{0.631}   & \textbf{0.478}   \\
\bottomrule
\end{tabular}%
\end{table*}

\section{Results and Analysis}

\subsection{Overall Results}

The comparative results with baseline models are presented in Table \ref{tab:baselines}, revealing several key observations.
First, we find that the TSF+Text achieves a lower average prediction error than the time-series-only baseline (TSF-only), whereas the improvement lacks uniformity across different domains. 
This finding indicates that while auxiliary textual data potentially enhances TSF precision, the benefits remain constrained by domain-specific text quality variations. 
Consequently, textual augmentation fails to deliver consistent superior performance universally. 
Second, TSF complemented solely by regenerated text from the LLM without RL (TSF+TeR) yields a higher error than both TSF+Text and TSF-only.
This outcome reveals that LLM-generated text, prompted without specific guidance, often introduces content irrelevant to the forecasting objectives or omits crucial information present in the original data. 
Such degradation in text utility directly undermines forecasting effectiveness. 
Third, integrating text reinforced by prediction-accuracy-optimized LLMs (TSF+TeR+r1) consistently outperforms TSF-only, TSF+Text, and TSF+TeR. 
This demonstrates that reinforcement learning directed by prediction accuracy rewards enables LLMs to generate text truly beneficial for time series forecasts, surpassing the utility of raw textual input. 
Finally, the framework combining dual-reward reinforced text (TSF+TeR+r12), utilizing both prediction accuracy and task relevance rewards, achieves the minimal prediction error overall. 
This enhancement underscores that the supplementary task relevance reward effectively guides LLM output towards lexically precise and contextually appropriate text generation. 
The stricter linguistic refinement consequently further elevates text quality, propelling forecasting performance beyond all other tested approaches. 
Collective evidence identifies text quality, ensured by reward-structured LLM guidance, as the key determinant of successful multimodal TSF enhancement.

We further compare the performance of existing state-of-the-art multimodal TSF models against the proposed TeR-TSF. The experiment results are summarized in Table \ref{tab:comp_sota}.
We observe that TeR-TSF consistently achieves the lowest or second-lowest prediction error across each domain tested while securing the minimal overall average error. 
This performance confirms that TeR-TSF delivers the highest prediction accuracy available, surpassing current state-of-the-art effectiveness. 
Particularly noteworthy, TeR-TSF reports consistently reduced errors compared to models (e.g., TFHTS \cite{textfusionhts}, Time-LLM \cite{time-llm}, and MM-TSF \cite{time-mmd}) that incorporate textual data without specific optimization processes for text quality.
This comparison underscores the effectiveness of TeR-TSF's text quality enhancement strategy. 
Optimizing text relevance and information content minimizes internal noise interference, enabling the multimodal model to extract genuinely useful knowledge from the textual input. 
An alternate approach, represented by MCD-TSF \cite{mcd-tsf}, attempts to mitigate textual interference by controlling the influence weight assigned to text during the forecasting process. 
Although this mechanism achieves lower error rates than TeR-TSF in three specific domains, the deliberate reduction of textual influence inherently limits MCD-TSF’s ability to exploit key information present within high-quality text. 
Consequently, its forecasting capability exhibits inherent limitations relative to text-optimized frameworks under conditions where valuable textual information exists. 
This comparative evidence collectively validates that TeR-TSF achieves superior multimodal TSF performance by fundamentally elevating text input utility, explaining its attainment of state-of-the-art prediction accuracy.

\subsection{Effect of Generated Reinforced Text Number}

\begin{figure}[t]
    \centering
    \includegraphics[width=\linewidth]{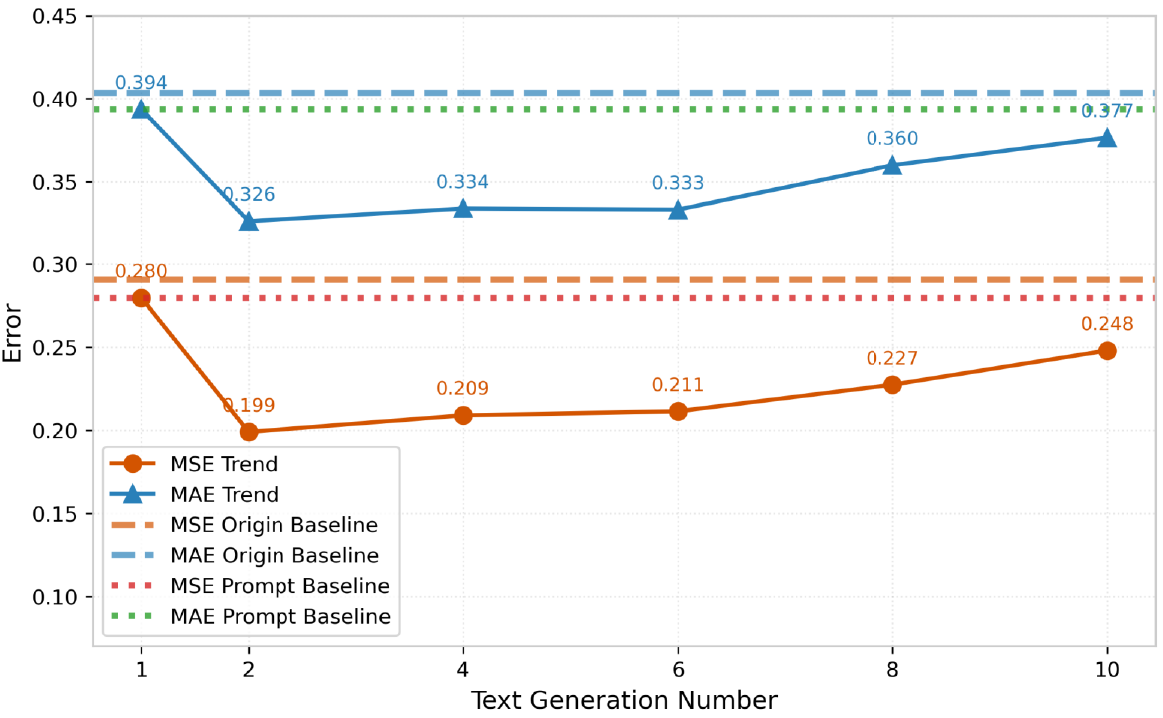}
    \caption{{Forecasting errors with different numbers of reinforced text number $k=\{1,2,4,6,8,10\}$ on the Energy domain. Solid lines indicate the variations of MSE and MAE with respect to $k$. Dashed lines denote the performance of the origin baseline and the prompt baseline, where the origin baseline refers to the multimodal TSF model with original text, and the prompt baseline corresponds to the multimodal TSF model using generated text without RL.}}
    \label{fig: text number}
\end{figure}

{
To investigate the impact of the number of generated augmented texts, denoted as $k$, used per reinforcement learning training iteration on model performance, we conduct experiments with varying values of $k$. 
Specifically, we test different settings for $k$, including $k=1, 2, 4, 6, 8$, and $10$. Fig. \ref{fig: text number} presents the resulting MSE and MAE curves for these configurations evaluated on the Energy dataset. 
The case where $k=1$ corresponds to using only the prompt-based approach without additional DPO training. 
Analysis of the results reveals that the lowest MSE and MAE values occur when $k$ increases from $1$ to $2$. 
This improvement demonstrates that introducing pairs of positive and negative text examples through DPO training enables the model to learn effective patterns within the textual guidance, consequently enhancing the performance of multimodal time series forecasting. 
However, further increasing $k$ beyond $2$ leads to a degradation in model performance, evidenced by rising error curves. 
We attribute this performance drop to the increased generation of lower-quality text as the total number of generated texts $k$ grows. 
This increase introduces more low-quality text into the training process. Consequently, the preference pairs formed exhibit excessively extreme quality differences.
This situation causes the model to ignore examples of intermediate quality during DPO training. 
Consequently, the DPO process encounters only extreme contrastive cases, lacking exposure to gradual learning examples that represent finer quality distinctions. 
This absence of progressive learning signals forces the model to learn with excessively large steps, causing the acquisition of suboptimal patterns. Therefore, we select the setting $k=2$ for our final approach. 
This configuration guarantees clear and informative learning signals during each training iteration and thus leads to better performance.
}

\begin{figure}[t]
    \centering
    \includegraphics[width=\linewidth, trim=0 0 0 0]{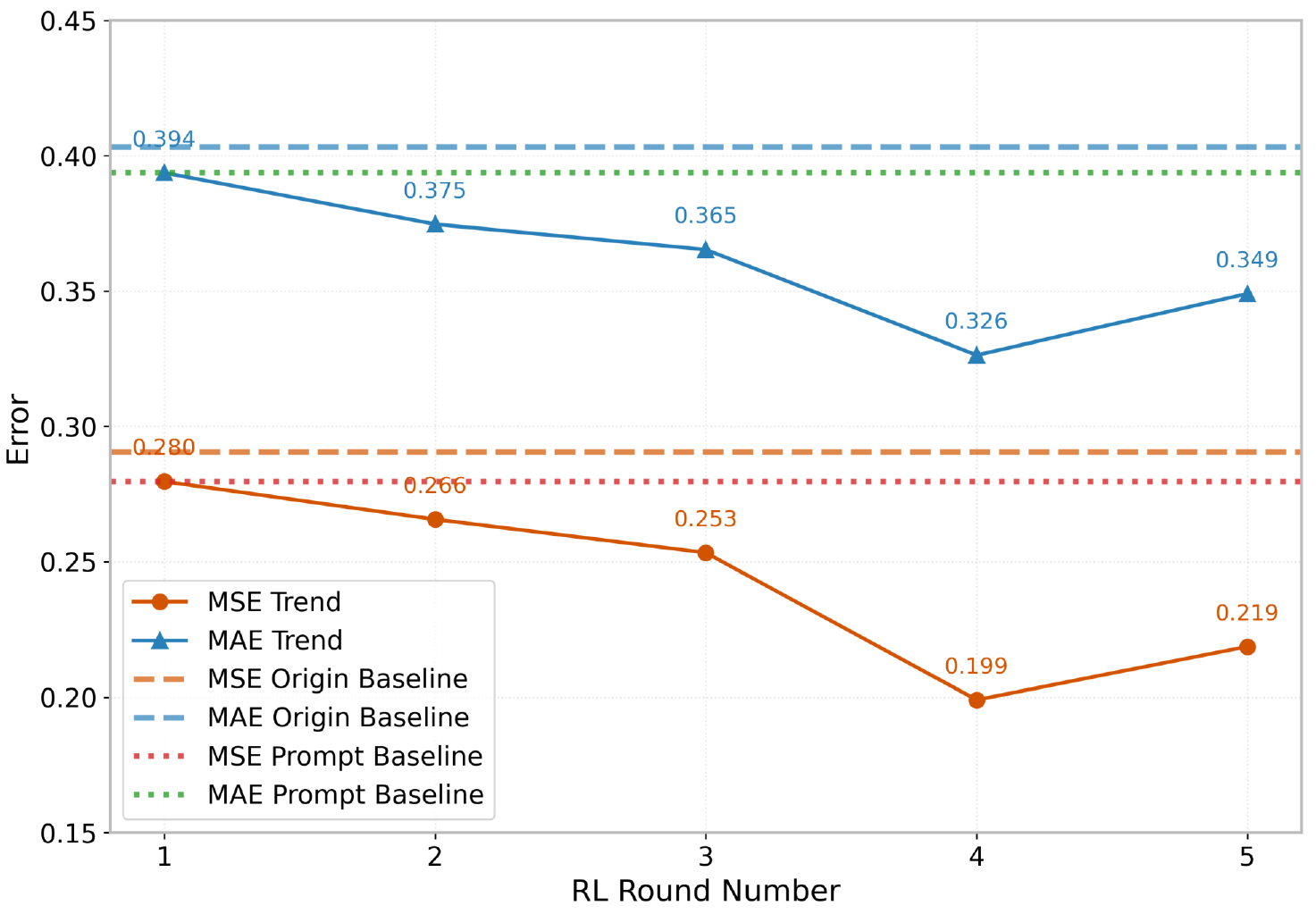}
    \caption{{Forecasting errors with different numbers of RL round $m=\{1,2,3,4,5\}$ on the test set of Energy domain. Solid lines indicate the variations of MSE and MAE with respect to $m$. Dashed lines denote the results of the origin baseline and the prompt baseline, where the origin baseline refers to the multimodal TSF model with the original text, and the prompt baseline corresponds to the multimodal TSF model using generated text without RL.}}
    \label{fig:rl round}
    \vspace{-0.1cm}
\end{figure}

\begin{figure}[t]
    \centering
    \includegraphics[width=\linewidth]{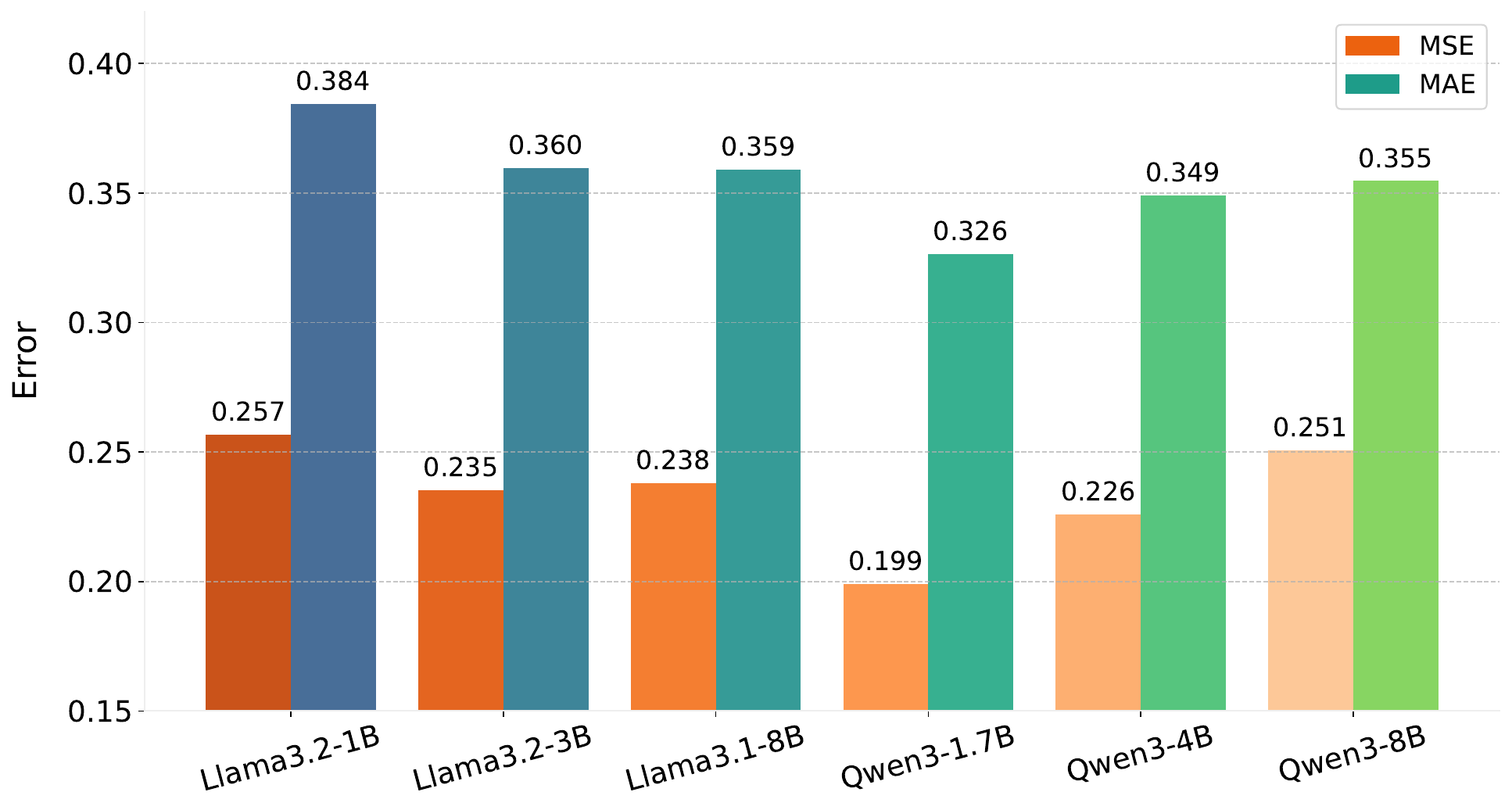}
    \caption{{Forecasting errors with different LLMs on the Energy dataset. The evaluated LLMs include the Llama series with parameter sizes of 1B, 3B, and 8B, and the Qwen series with parameter sizes of 1.7B, 4B, and 8B. For every model configuration, both MSE and MAE are reported.}}
    \label{fig:llm type}
\end{figure}

\begin{figure}
    \centering
    \includegraphics[width=\linewidth, trim=0 3 0 0]{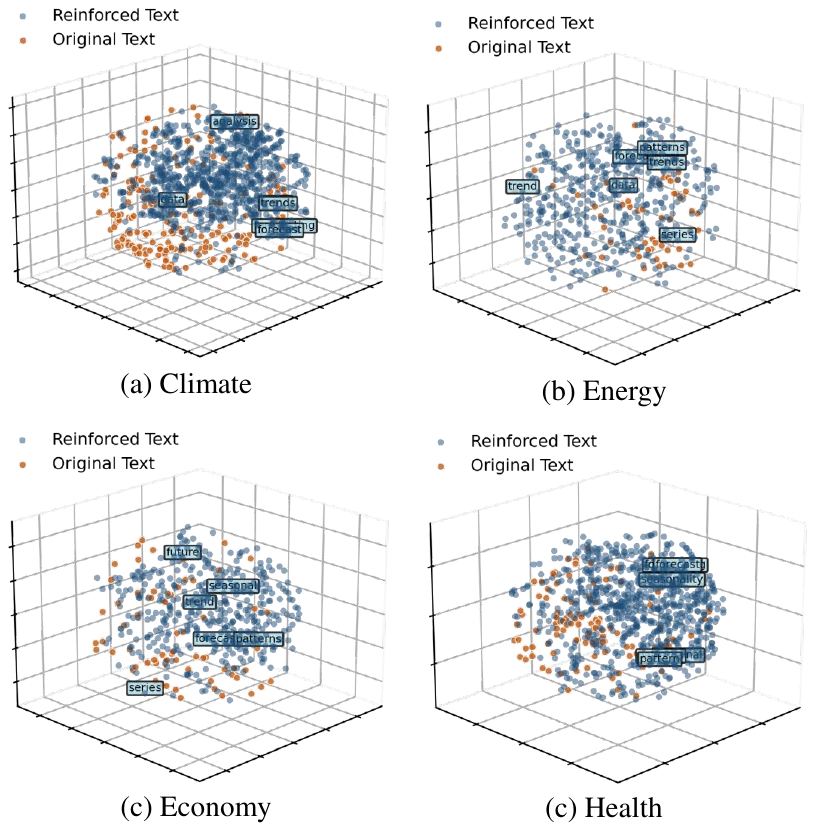}
    \caption{{t-SNE visualization of the vocabularies from original text and reinforced text. The plots are drawn separately for the four domains, namely, Climate, Energy, Economy, and Health. In each domain, representative tokens from the reinforced text that show high relevance to time series forecasting tasks are highlighted.}}
    \label{fig: t-sne}
    \vspace{-0.1cm}
\end{figure}

\begin{figure*}[t]
    \centering
    \includegraphics[width=\textwidth]{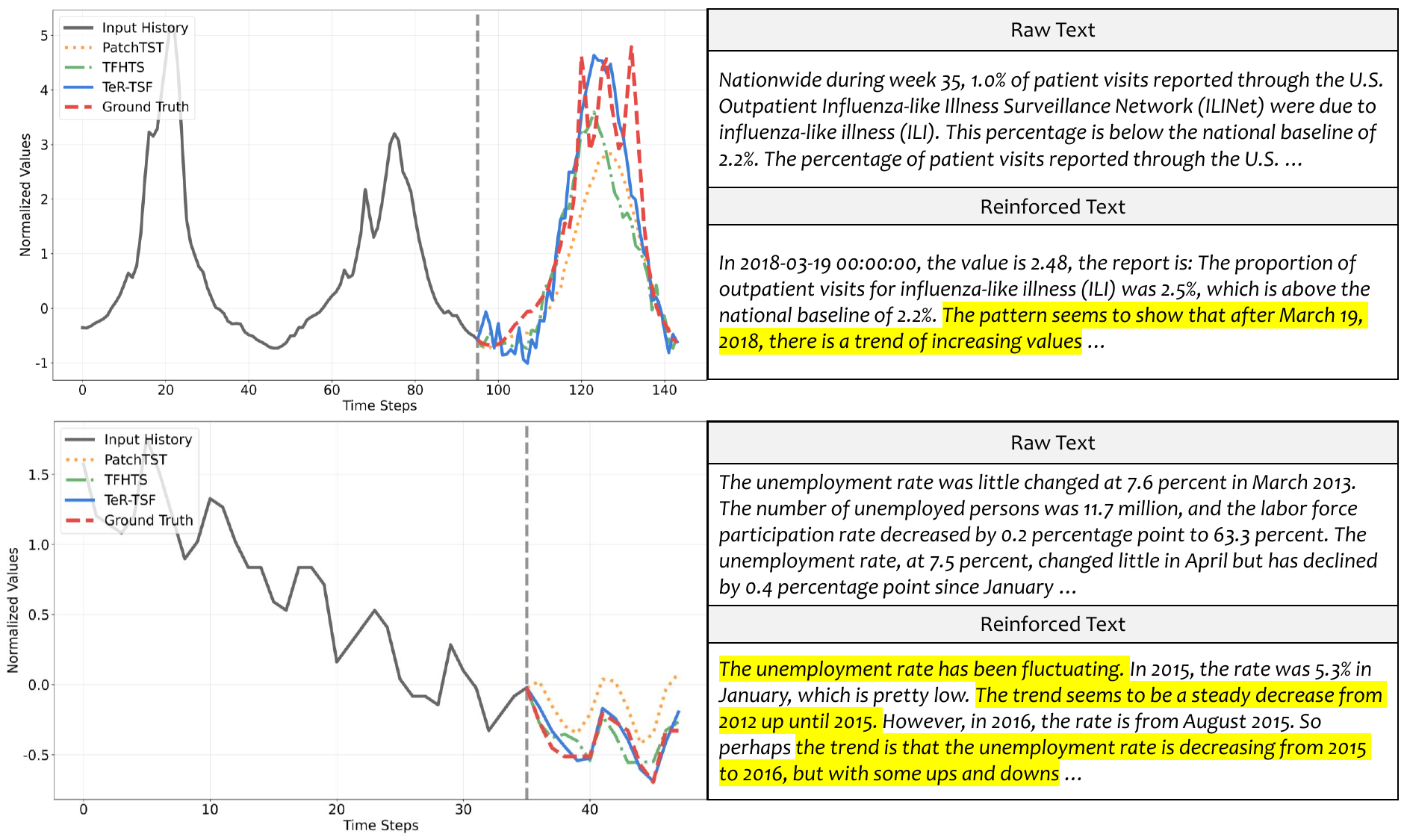}
    \caption{{
    Case study with two examples from Health (top panels) and SocialGood (bottom panels) domains. The left side compares model forecasts where the black line denotes the historical time series, the red dashed line marks the ground truth, and gray vertical dashed lines separate the historical window from the prediction window. Within the prediction window, the orange dashed line reports the PatchTST result as a representative unimodal predictor, the green dashed-dotted line reports the TFHTS result, and the blue solid line reports the TeR-TSF result. The right side lists the Raw Text used as the TFHTS input and the Reinforced Text produced by the TeR module and used as the TeR-TSF input, with phrases that describe temporal variation patterns highlighted with a yellow background.
    }}
    \label{fig: case study}
\end{figure*}

\subsection{Effect of RL Rounds}
{
To investigate the impact of the number of RL training rounds (denoted as $m$) on model performance, we conduct experiments by training the model with varying $m$ values ranging from $1$ to $5$. The results are visualized in Fig. \ref{fig:rl round}. A higher $m$ implies more iterative refinements during optimization.
There are some key observations.
First, the MSE and MAE curves initially decrease as $m$ increases, reaching their lowest values at $m = 4$. 
This demonstrates that an appropriate number of training rounds enhances model performance. 
The performance deficit at lower $m$ values ($m<4$) stems from insufficient training, resulting in an unrefined text augmentation strategy that demonstrates underfitting.
At $m=4$, the model achieves its optimal learning state, effectively capturing the correct textual preferences necessary for high-quality augmentation. 
However, increasing $m$ beyond this point ($m>4$) leads to a rise in both the MSE and MAE curves. 
This reversal demonstrates that an excessive number of training rounds degrades model performance. 
The degradation occurs because excessive training causes the model to overfit the limited training data. 
Consequently, the learned strategy for generating augmented text loses its ability to generalize effectively to data not encountered during the training process. 
Given these findings, we set $m=4$ for all subsequent experiments. 
This specific configuration achieves the optimal balance point where the model learns sufficiently without succumbing to overfitting, maximizing its ability to generalize while minimizing prediction errors on the evaluation metrics.
}

\subsection{Effect of Text Enhanced LLM}

{
To investigate the impact of different LLMs serving as the TeR module, we evaluate models from the Llama series \cite{dubey2024llama} (Llama3.2-1B, Llama3.2-3B, Llama3.1-8B) and the Qwen series \cite{yang2025qwen3} (Qwen3-1.7B, Qwen3-4B, Qwen3-8B). 
The comparative results are presented in Fig. \ref{fig:llm type}. 
Key observations emerge from this analysis. 
For the Llama models, increasing the model size to 3B or 8B improves prediction accuracy, indicated by a reduction in MSE and MAE.
Performance between Llama3.2-3B and Llama3.1-8B is comparable.
This observation implies that the Llama architecture reaches a plateau in prediction accuracy at the 3B scale.
Further parameter increases do not yield additional gains.
The analysis proposes that time series datasets generally possess a much smaller scale than the text corpora used for LLM pre-training, causing overfitting with excessively large parameter counts.
For the Qwen models, the 1.7B parameter version delivers the top performance, marked by the lowest errors.
Expanding the model size to 4B or 8B parameters fails to improve performance accordingly, indicating that the Qwen series achieves performance saturation at the 1.7B scale.
The difference between Llama and Qwen outcomes stems from distinctions in model architecture and pre-training procedures between these two LLM families.
The analysis demonstrates that the optimal LLM size for the TeR component depends on the specific task and data characteristics. 
Smaller models such as Qwen3-1.7B prove most effective here by balancing task capability with robustness against overfitting under dataset size constraints.
}

{
To further investigate the role of the TeR in reinforcing text, we present t-SNE visualizations in Fig. \ref{fig: t-sne} illustrating the distribution of tokens from the original text compared to tokens within the reinforced text in four domains. 
This visualization aims to understand the transformation of token representations induced by the LLM's reinforcement process. 
We observe that the tokens from the original text and those from the reinforced text consistently form two distinct clusters across all domains. 
This separation indicates a distribution gap between tokens in the original text and tokens in the reinforced text. 
The finding demonstrates that the TeR transforms the original text into a new distribution. 
Furthermore, this new distribution contains substantially more terms directly related to temporal patterns in time series data.
For instance, enhanced text frequently incorporates tokens such as ``\textit{trend}'' and ``\textit{seasonality}'', which describe core characteristics of time series fluctuations.
This evidence confirms that our proposed approach effectively filters out irrelevant tokens from the original text while generating new text highly relevant to time series forecasting tasks.
The generated text possesses greater utility for multimodal time series forecasting models, as it focuses explicitly on temporally informative concepts.
The consistent cluster separation across diverse domains highlights the generalization capability of the text enhancement process. 
Tokens associated with temporal dynamics become predominant in the enhanced representations, replacing less relevant vocabulary from the original descriptions. 
This redistribution of semantic focus underpins the performance gains observed when using enhanced text. 
The analysis provides direct empirical support for the mechanism through which TeR improves forecasting accuracy by concentrating linguistic information on time-dependent patterns essential for prediction. 
The generated text aligns more closely with the underlying structures that the forecasting model captures.
}

\subsection{Case Study}
{
To qualitatively validate the effectiveness of our proposed approach, Fig. \ref{fig: case study} displays two examples, one from the Health domain (upper panel) and one from the SocialGood domain (lower panel), comparing predictions from PatchTST \cite{patchtst}, TFHTS \cite{textfusionhts}, and TeR-TSF.
The figure’s right part presents the raw text input for TFHTS and the reinforced text input for TeR-TSF.
The phrases that describe the temporal variation patterns in the text generated by our approach are highlighted in a yellow background.
The case study indicates that TFHTS, the multimodal forecasting model using raw text, generates a prediction curve aligning more precisely with Ground Truth compared to the unimodal PatchTST model.
This demonstrates the benefit of integrating textual information for forecasting. The text modality potentially improves prediction accuracy, supported by the raw text content, which provides partial descriptions of historical states.
Further comparison shows TeR-TSF, utilizing reinforced text, provides predictions with greater accuracy than TFHTS. TeR-TSF’s curve displays closer alignment with Ground Truth, an outcome supported by analysis of the reinforced text.
The reinforced text contains extensive analysis elucidating historical time series change patterns. These analytical descriptions prove more useful for the multimodal forecasting model than the raw text, which solely describes past states without interpreting temporal dynamics.
Collectively, these findings confirm that our approach generates high-quality enhanced text and reliably improves prediction accuracy in multimodal time series forecasting.
}

\section{Conclusion}
{
This paper presents a multimodal time series data reinforcement approach to enhance the performance of multimodal TSF models.
The approach utilizes LLMs, taking both the original time series data and any available raw text as input, to generate reinforced textual data.
These reinforced texts are then concatenated with the original time series to form enriched multimodal inputs, enabling the downstream forecasting model to leverage enhanced semantic clues and capture complex temporal patterns more effectively.
To train the text reinforcement model, we employ a reinforcement learning pipeline, where the LLM generates multiple candidate texts that are scored by a multi-dimensional reward function at each iteration.
Preference pairs derived from the highest- and lowest-scoring candidates are used by DPO to iteratively optimize the LLM’s generation strategy, providing richer contextual information for multimodal TSF models and leading to more informed predictions.
Extensive experiments on a real-world benchmark dataset demonstrate the effectiveness of our approach, showing consistent improvements in prediction accuracy. 
Analyses further confirm the contribution of each proposed module. 
}

\nocite{langley00}

\ifCLASSOPTIONcaptionsoff
  \newpage
\fi

\bibliographystyle{IEEEtran}
\bibliography{my_paper}

\newpage

\appendices

\end{document}